\documentclass[a4paper,fleqn]{cas-sc}

\usepackage[numbers]{natbib}
\def\tsc#1{\csdef{#1}{\textsc{\lowercase{#1}}\xspace}}
\tsc{WGM}
\tsc{QE}
\tsc{EP}
\tsc{PMS}
\tsc{BEC}
\tsc{DE}
\usepackage{epigraph}
\usepackage{longtable}
\usepackage{setspace}

\usepackage{textcomp}

\usepackage{multirow}

\usepackage{color}
\definecolor{WordGreen}{RGB}{100, 136, 40}
\definecolor{WordDarkGrey}{RGB}{82, 82, 82}
\definecolor{WordRed}{RGB}{192, 80, 77}
\definecolor{WordBlue}{RGB}{0, 122, 192}

\definecolor{WordLightBlue}{RGB}{218, 238, 243}
\definecolor{WordLightGreen}{RGB}{234, 241, 221}

\definecolor{WordFillGreen}{RGB}{194, 214, 155}
\definecolor{WordFillRed}{RGB}{252, 214, 182}
\definecolor{WordFillGray}{RGB}{217, 217, 217}

\UseRawInputEncoding
\usepackage{times}
\usepackage{epsfig}
\usepackage{amssymb}
\usepackage[ruled,vlined,linesnumbered]{algorithm2e}
\usepackage{comment}
\usepackage{array, makecell} 
\usepackage{lettrine}
\usepackage{url} 
\usepackage{lscape}
\usepackage{cuted}
\usepackage{tabularx}
\usepackage{colortbl}
\usepackage[normalem]{ulem}
\usepackage{hhline}
\usepackage{vcell}
\usepackage{longtable}
\usepackage{vcell}
\usepackage{rotating}
\usepackage{pdflscape}
\usepackage{sidecap}
\usepackage{neuralnetwork}
\usepackage[export]{adjustbox} % https://tex.stackexchange.com/questions/91566/syntax-similar-to-centering-for-right-and-left
\usepackage[accsupp]{axessibility} % Improves PDF readability for those with disabilities.
\usepackage{amsfonts,bm}
\usepackage{physics}
\usepackage{wrapfig}
\usepackage{caption}
\usepackage{color,soul}
\usepackage[frozencache,cachedir=minted-cache]{minted}
\usepackage{mathtools}
\usepackage{amsthm}

\usepackage{acronym}
\acrodef{FCN}[FCN]{Fully Convolutional Network}
\acrodef{GAME}[GAME]{Grid Average Mean Absolute Error}
\acrodef{DL}[DL]{Deep Learning}
\acrodef{DNN}[DNN]{Deep Neural Network}
\acrodef{ML}[ML]{Machine Learning}
\acrodef{CV}[CV]{Computer Vision}
\acrodef{AI}[AI]{Artificial Intelligence}
\acrodef{CNN}[CNN]{Convolutional Neural Network}
\acrodef{RNN}[RNN]{Recurrent Neural Network}
\acrodef{GAN}[GAN]{Generative Adversarial Network}
\acrodef{JCU}[JCU]{James Cook University}
\acrodef{MAE}[MAE]{Mean Average Error}
\acrodef{MAP}[mAP]{Mean Average Precision}
\acrodef{CA}[CA]{Classification Accuracy}
\acrodef{LCFCN}[LCFCN]{Localization-based Counting loss Fully Convolutional Network}
\acrodef{IoT}[IoT]{Internet of Things}
\acrodef{MLP}[MLP]{Multi-Layer Perceptrons}
\def\q{{\mathbf q}}				% Definition of the bold letter q
\usepackage{xspace}

\usepackage[capitalize]{cleveref}
\crefname{section}{Sec.}{Secs.}
\Crefname{section}{Section}{Sections}
\Crefname{table}{Table}{Tables}
\crefname{table}{Table}{Table}
\usepackage{booktabs}
\usepackage{arydshln}
\usepackage{listings}
\definecolor{codegreen}{rgb}{0,0.6,0}
\definecolor{codegray}{rgb}{0.5,0.5,0.5}
\definecolor{codepurple}{rgb}{0.58,0,0.82}
\definecolor{backcolour}{rgb}{0.95,0.95,0.92}
\definecolor{cleacolorr}{rgb}{1,1,1}

\lstdefinestyle{mystyle}{
    backgroundcolor=\color{cleacolorr},   
    commentstyle=\color{codegreen},
    keywordstyle=\color{magenta},
    numberstyle=\tiny\color{codegray},
    stringstyle=\color{codepurple},
    basicstyle=\ttfamily\footnotesize,
    breakatwhitespace=false,         
    breaklines=true,                 
    captionpos=b,                    
    keepspaces=true,                 
    numbers=left,                    
    numbersep=5pt,                  
    showspaces=false,                
    showstringspaces=false,
    showtabs=false,                  
    tabsize=2
}

\usepackage[most]{tcolorbox}
\newtcolorbox[auto counter]{pabox}[2][]{%
colback=blue!5!white,colframe=blue!75!black,fonttitle=\bfseries,
title=Box~\thetcbcounter: #2,#1}
\usepackage{tikz}
\usepackage{graphicx}
\graphicspath{{./Graphics/}}
\DeclareGraphicsExtensions{.pdf,.jpeg,.png,.jpg}

\usepackage{amsmath}

\usepackage{array}

\usepackage{url}
\begin{document}
% \doublespacing
% \linenumbers

\let\WriteBookmarks\relax
\def\floatpagepagefraction{1}
\def\textpagefraction{.001}

\shorttitle{Semi-Supervised Weed Detection with Multi-scale Representation}

\shortauthors{Saleh et~al.}

\title [mode = title]{%Shadow Removal Algorithms Trained On Unpaired Data For Use In Field Robotics
%Semi-Supervised Weed Detection with Multi-scale Representation
%RT-SRNet: Real-Time Shadow Removal Network for Edge and Robotics
Semi-Supervised Weed Detection for Rapid Deployment and Enhanced Efficiency
}                      
% \tnotemark[1,2]

% \tnotetext[1]{This document is the results of the research
%    project funded by the National Science Foundation.}

% \tnotetext[2]{The second title footnote which is a longer text matter
%    to fill through the whole text width and overflow into
%    another line in the footnotes area of the first page.}

\author[1]{Alzayat Saleh}[orcid=0000-0001-6973-019X]
\credit{Conceptualisation, Data Curation, Data Analysis, Software Development, DL Algorithm Design, Visualization, Writing original draft}

% % Email id of the first author
% \ead{alzayat.saleh@my.jcu.edu.au}

\author[2]{Alex Olsen}
% [                        orcid=0000-0001-7511-2910]
\credit{Data Curation, Data Analysis, Reviewing/editing the draft}

\author[2]{Jake Wood}
% [                        orcid=0000-0001-7511-2910]
\credit{Data Curation, Data Analysis}

\author[1]{Bronson Philippa}[orcid=0000-0002-5736-0336]
\credit{Reviewing/editing the draft}

\author[1,3]{Mostafa~Rahimi~Azghadi}[                        orcid=0000-0001-7975-3985]
\cormark[1]
\ead{mostafa.rahimiazghadi@jcu.edu.au}
\credit{Conceptualisation, Data Curation, Data Analysis, Supervision, Reviewing/editing the draft}
% \author[1,3]{Alzayat Saleh}[                        orcid=0000-0001-7511-2910]
% \author[1,3]{Alzayat Saleh}[                        orcid=0000-0001-7511-2910]

% % Corresponding author indication
% \cormark[1]

% % Footnote of the first author

% % Email id of the first author
% \ead{alzayat.saleh@my.jcu.edu.au}

% % URL of the first author
% \ead[url]{www.cvr.cc, cvr@sayahna.org}

\affiliation[1]{organization={College of Science and Engineering, James Cook University},
    % addressline={Radarweg 29}, 
    city={Townsville},
    % citysep={}, % Uncomment if no comma needed between city and postcode
    postcode={4814}, 
    state={QLD},
    country={Australia}}

% % Second author
% \author[2,4]{Han Theh Thanh}[style=chinese]

% % Third author
% \author[2,3]{CV Rajagopal}[%
%    role=Co-ordinator,
%    suffix=Jr,
% \ead{cvr3@sayahna.org}
% \ead[URL]{www.sayahna.org}

% \credit{Data curation, Writing - Original draft preparation}

\affiliation[2]{organization={AutoWeed Pty Ltd},
    city={Townsville},
    % citysep={}, % Uncomment if no comma needed between city and postcode
    postcode={4814}, 
    state={QLD},
    country={Australia}}

% % Fourth author
% \author%
% [1,3]
% {Rishi T.}
% \cormark[2]
% \fnmark[1,3]
% \ead{rishi@stmdocs.in}
% \ead[URL]{www.stmdocs.in}

% \affiliation[3]{organization={STM Document Engineering Pvt Ltd.},
%     addressline={Mepukada}, 
%     city={Malayinkil},
%     % citysep={}, % Uncomment if no comma needed between city and postcode
%     postcode={695571}, 
%     state={Trivandrum},
%     country={India}}

\affiliation[3]{organization={Agriculture Technology and Adoption Centre, James Cook University},
    % addressline={Radarweg 29}, 
    city={Townsville},
    % citysep={}, % Uncomment if no comma needed between city and postcode
    postcode={4814}, 
    state={QLD},
    country={Australia}}

\cortext[cor1]{Corresponding author}
% \cortext[cor2]{Principal corresponding author}

% % Footnote text
%\fntext[fn1]{This is the first author footnote.}
% \fntext[fn2]{Another author footnote, this is a very long footnote and
%   it should be a really long footnote. But this footnote is not yet
%   sufficiently long enough to make two lines of footnote text.}

% % For a title note without a number/mark
% \nonumnote{This note has no numbers. In this work we demonstrate $a_b$
%   the formation Y\_1 of a new type of polariton on the interface
%   between a cuprous oxide slab and a polystyrene micro-sphere placed

\begin{abstract}
Weeds present a significant challenge in agriculture, causing yield loss and requiring expensive control measures. Automatic weed detection using computer vision and deep learning offers a promising solution. However, conventional deep learning methods often require large amounts of labelled training data, which can be costly and time-consuming to acquire.
This paper introduces a novel method for semi-supervised weed detection, comprising two main components. Firstly, a multi-scale feature representation technique is employed to capture distinctive weed features across different scales. Secondly, we propose an adaptive pseudo-label assignment strategy, leveraging a small set of labelled images during training. This strategy dynamically assigns confidence scores to pseudo-labels generated from unlabeled data. Additionally, our approach integrates epoch-corresponding and mixed pseudo-labels to further enhance the learning process. 
%Our method relies only on a small amount of labelled data and effectively utilises unlabelled data for learning. We introduce a multi-scale feature representation technique within the deep learning model to capture discriminative features specific to weeds at different scales. Additionally, we employ an adaptive pseudo-label assignment strategy that dynamically assigns confidence scores to pseudo-labels generated from unlabeled data during training. This approach, combined with epoch-corresponding and mixed pseudo-labels, further improves the quality of the learning process.
Experimental results on the COCO dataset and five prominent weed datasets -- CottonWeedDet12, CropAndWeed, Palmer amaranth, RadishWheat, and RoboWeedMap -- illustrate that our method achieves state-of-the-art performance in weed detection, even with significantly less labelled data compared to existing techniques. This approach holds the potential to alleviate the labelling burden and enhance the feasibility and deployment speed of deep learning for weed detection in real-world agricultural scenarios.
\end{abstract}

% \includegraphics{figs/grabs.pdf}

% % Research highlights
% \begin{highlights}
% \item Research highlights item 1
% \item Research highlights item 2
% \item Research highlights item 3
% \end{highlights}

% Keywords

\maketitle

\section{Introduction}\label{secintro}

Weeds are a major agricultural concern, causing significant yield losses and demanding extensive control measures. Traditional methods like manual removal or herbicide application are labour-intensive, expensive, and may impair the ecosystem.
Automated weed detection using computer vision offers a promising alternative. Vision-based systems provide several advantages:
1) Efficiency: They significantly reduce the time and labour required for weed identification and control.
2) Precision: They can potentially achieve higher accuracy in weed identification compared to manual methods.
3) Scalability: They can be applied to large fields or farms, facilitating efficient weed management.
4) Sustainability: They can provide valuable environmental benefits by reducing the amount of herbicides for weed management through weed spot-spraying.

However, deep learning-based weed detection methods often face a significant hurdle: the need for vast amounts of labelled training data.  Labeling images for weed identification is time-consuming and expensive. Limited labelled data can lead to
1) Overfitting: Models trained with limited data struggle to generalize well to unseen images, resulting in inaccurate real-world performance.
2) Inefficient Learning: Traditional models might not fully utilize the information contained in unlabeled data, which could be valuable for improving weed detection capabilities.

 \begin{figure}
    \centering
    \includegraphics[width=0.60\linewidth]{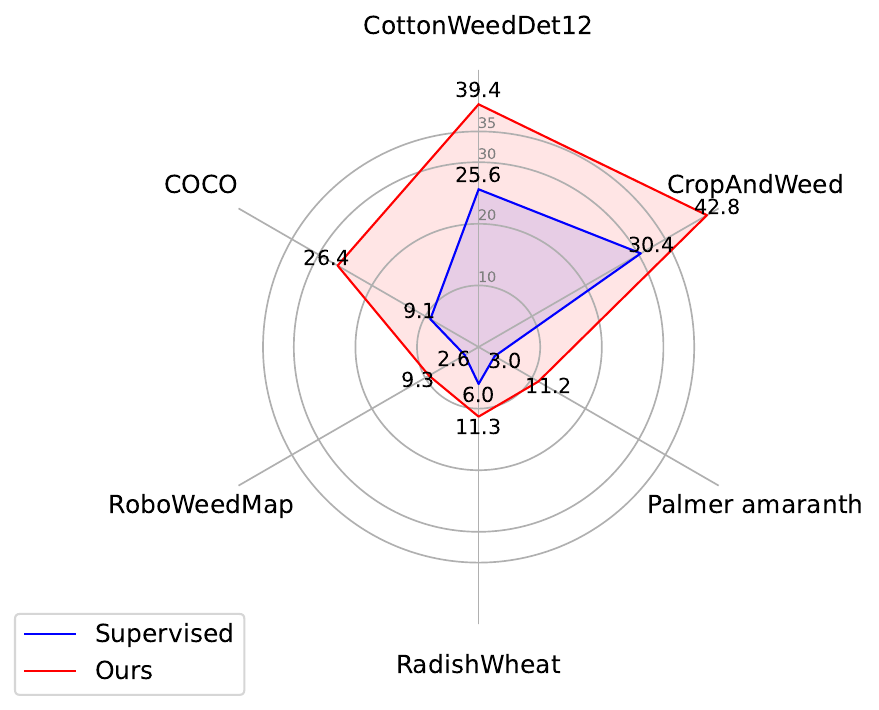}
    \caption{Performance of our proposed semi-supervised model compared to a supervised model, when trained on 1\% of available labelled training data across COCO and several weed datasets. See \cref{tab_1} and \cref{tab_2} for more details.}
    \label{l_fig_0}
\end{figure}

Semi-Supervised Object Detection (SSOD) techniques offer a promising approach to address these challenges. SSOD methods utilize unlabeled data alongside a smaller amount of labelled data to improve model performance and overcome limitations associated with data scarcity. However, applying SSOD to one-stage anchor-based detectors, a popular choice for their high recall, stability, and speed, requires further research due to specific challenges.
These challenges include:
1) Pseudo Label Inconsistency: During SSOD training, the teacher model generates pseudo labels for unlabeled data. However, the quality and quantity of these labels can fluctuate significantly, potentially misleading the student model. This "pseudo label inconsistency" is particularly problematic for one-stage anchor-based detectors due to their abundance of predictions from the multiple-anchor mechanism.
2) Accuracy vs. Efficiency Trade-off: Existing SSOD methods often prioritize high accuracy but sacrifice training efficiency. This limits their applicability to real-world scenarios, such as weed detection in agriculture, which needs fast, resource-efficient detectors with high accuracy.

This paper proposes a novel method for semi-supervised weed detection with multi-scale Representation specifically for one-stage anchor-based detectors. We aim to address the challenges of pseudo-label inconsistency and develop SSOD methods that achieve a balance between accuracy and efficiency, making them suitable for real-world agricultural applications.
By introducing a multi-scale Representation technique, we enhance our model's ability to learn informative representations from both labelled and unlabeled images. This technique allows the model to better distinguish weeds from other objects in the agricultural environment.
In essence, as shown in Figure~\ref{l_fig_0}, our proposed model achieves state-of-the-art performance in weed detection across various datasets, even when trained on only 1\% of the labelled data, which is significantly less than what is typically required by supervised learning methods. These results demonstrate the effectiveness of our approach in leveraging unlabeled data through multi-scale Representation.

Our proposed method offers several benefits:
\begin{itemize}
    \item Reduced Labeling Burden: By utilizing unlabeled data, our approach significantly reduces the need for large amounts of labelled data, making the deep learning model more practical and cost-effective for real-world deployments.
    %\item Multi-scale Representation Technique: We introduce a novel multi-scale Representation technique to enhance the model's ability to learn informative representations from both labelled and unlabeled images. This technique helps identify discriminative features specific to weeds, leading to improved detection performance.
    \item Improved Weed Detection Performance: Experiments demonstrate that our method achieves state-of-the-art performance in weed detection even with a limited amount of labelled data.
    \item Enhanced Efficiency for Weed Management: This approach has the potential to improve the efficiency and accuracy of weed detection in agricultural applications, leading to better resource management and reduced environmental impact.
\end{itemize}

The remainder of this paper is organized as follows: Section~\ref{sec:related_work} reviews related research in weed detection and semi-supervised learning. Section~\ref{sec:methodology} presents the details of our proposed method with multi-scale Representation. Section~\ref{sec:experiments} describes our experimental setup, dataset, and results, comparing our approach to traditional methods. Section~\ref{sec:discussion} discusses the results and limitations of our work. Finally, Section~\ref{sec:conclusion} concludes the paper and outlines future research directions.

 \begin{figure}
    \centering
    \includegraphics[width=0.90\linewidth]{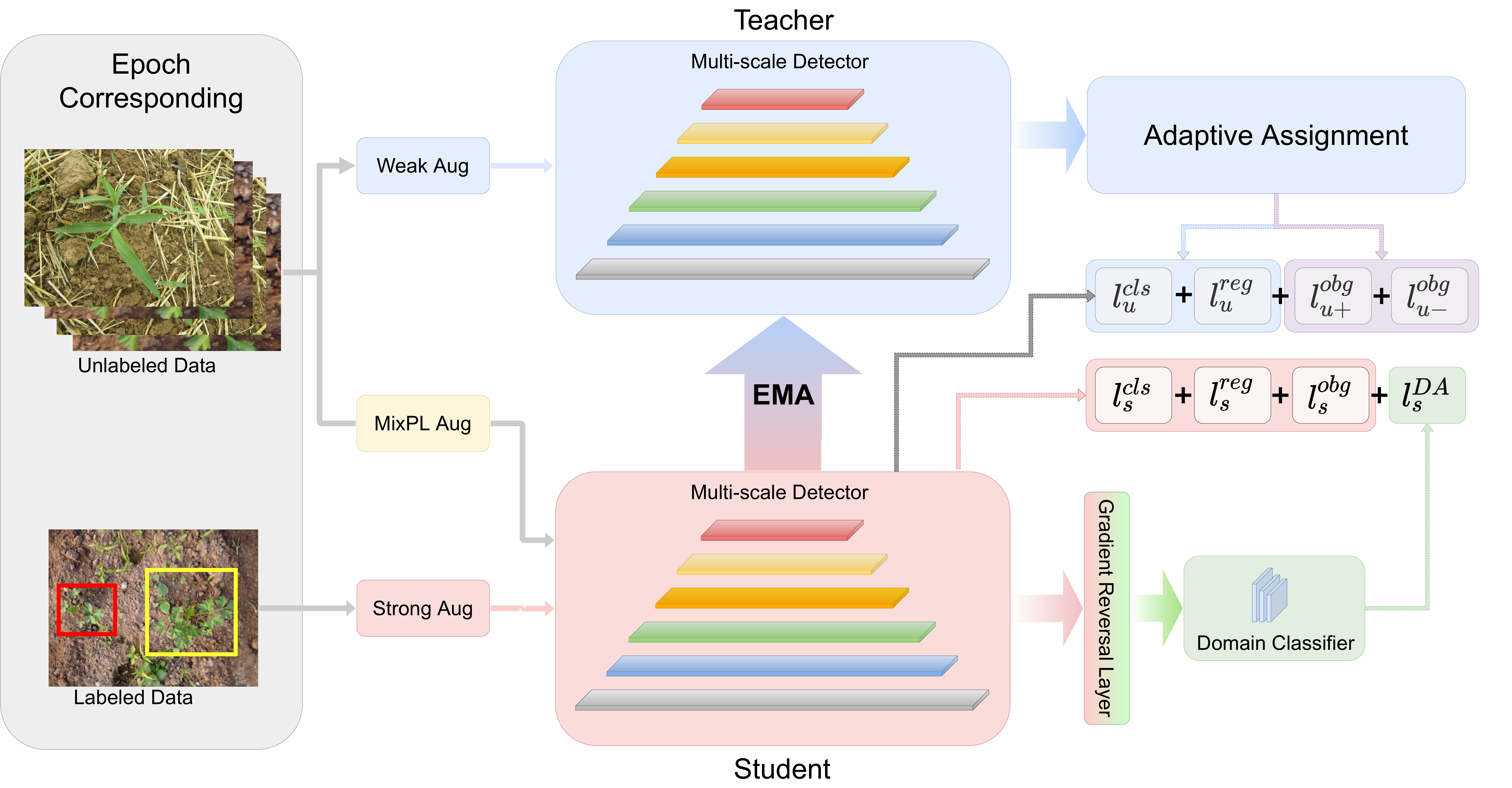}
    \caption{Multi-scale Detector}
    \label{l_fig_9}
\end{figure}

%%%%%%%%%%%%%%%%%%%%%%%%%%%%%%%%%%%%%%%%%%%%%%%%%%%%%%%%%%%%%%%%
 
\section{Related Work} \label{sec:related_work}

\textbf{Semi-supervised Object Detection,} is a paradigm that leverages the power of unlabeled data to enhance object detection performance, building upon the successes of semi-supervised learning in image classification.
There are two main approaches exists in SSOD:
a) Consistency Regularization: This approach enforces consistency between the model's predictions on the same unlabeled data under different transformations (e.g., rotations, flips)~\cite{Jeong2019Consistency-basedDetection,Tang2021ProposalDetection}.
b) Self-Training (Pseudo-labeling): This approach generates pseudo-labels for unlabeled data through a pre-trained model, then fine-tunes the model using both labeled and pseudo-labeled data~\cite{Sohn2020ADetection,Liu2021UnbiasedDetection,Xu2021End-to-endTeacher}. 

Recent research focuses on self-training methods, e.g. Mean Teacher.
Mean Teacher  method utilizes a moving average of the model's weights as a "teacher" for generating pseudo-labels, reducing the need for offline pseudo-labeling~\cite{Zhou2021Instant-teaching:Framework,Liu2021UnbiasedDetection,Xu2021End-to-endTeacher}.
Moreover, Focal Loss and Soft Weights  methods address foreground-background and class imbalance issues that are more prominent in object detection compared to classification~\cite{Liu2021UnbiasedDetection,Xu2021End-to-endTeacher}.
However, existing methods still struggle with label mismatch during pseudo-labeling. This paper proposes a novel approach to address this challenge and further improve SSOD performance.

\textbf{Accurate Label Assignment, }
which involves determining the classification and localization targets for each proposal or anchor, is crucial for achieving optimal object detection performance~\cite{Ge2021Ota:Detection}. The existing label assignment methods can be broadly categorized into two types: fixed and dynamic assignment.

Fixed assignment strategies, widely employed in both RCNN-series~\cite{Girshick2014, girshick2015fast, Ren2015FasterNetworksb, Dai2015, Pang2018} and one-stage detectors~\cite{Redmon2016YouDetection, Redmon2018Yolov3:Improvementb, lin2017focal, Liu2016}, encompass IoU-based and center-based approaches. Anchor-free detectors often rely on center-based assignment~\cite{Tian2019Fcos:Detection, Kong2020Cycle-contrastLearning}. Recent studies, including ATSS~\cite{Zhang2020BridgingSelection}, PAA~\cite{Kim2020ProbabilisticDetection}, AutoAssign~\cite{Zhu2020Autoassign:Detection}, and OTA~\cite{Ge2021Ota:Detection}, have proposed adaptive mechanisms for supervised object detection. However, Semi-Supervised Object Detection (SSOD) poses a unique challenge due to its inherent complexity.

This paper addresses the instance-level label mismatch problem in SSOD by proposing a novel label assignment method tailored specifically for self-training in one-stage anchor-based detectors. Existing supervised label assignment strategies often prove inadequate for SSOD, leading to performance degradation~\cite{Chen2022LabelDetection, Liu2022UnbiasedDetectors}. Our proposed method aims to overcome these limitations and improve SSOD performance.

\textbf{Multi-Scale Features,}
a technique proven effective since the early days of feature engineering~\cite{Dollar2014FastDetection}, is a common practice in CNNs for achieving scale-invariant object recognition~\cite{Wang2020DeepRecognition}. While recent vision transformers (ViTs) have explored multi-scale approaches~\cite{Yang2021FocalTransformers}, these typically require specialized architectures.

Scaling vision models by increasing parameters is the dominant pre-training approach~\cite{He2016DeepRecognition}. Research has focused on balancing model width, depth, and resolution to achieve optimal performance~\cite{Tan2019Efficientnet:Networks}. Although recent work has explored scaling ViTs~\cite{Cherti2023ReproducibleLearning}, incorporating high-resolution images during pre-training is often limited by computational resources~\cite{Oquab2023DINOv2:Supervision}.

This work addresses these limitations by proposing a novel approach that leverages both multi-scale representation and vision model scaling for improved semi-supervised weed detection. Our method enhances the model's ability to learn informative representations from both labeled and unlabeled data, leading to improved performance.

%%%%%%%%%%%%%%%%%%%%%%%%%%%%%%%%%%%%%%%%%%%%%%%%%%%%%%%%%%%%%%%%
 
\section{Method} \label{sec:methodology}

Our proposed method for semi-supervised weed detection with multi-scale Representation consists of three key components, which are described in detail in this section.

\begin{figure}[!t]
    \centering
    \includegraphics[width=0.90\linewidth]{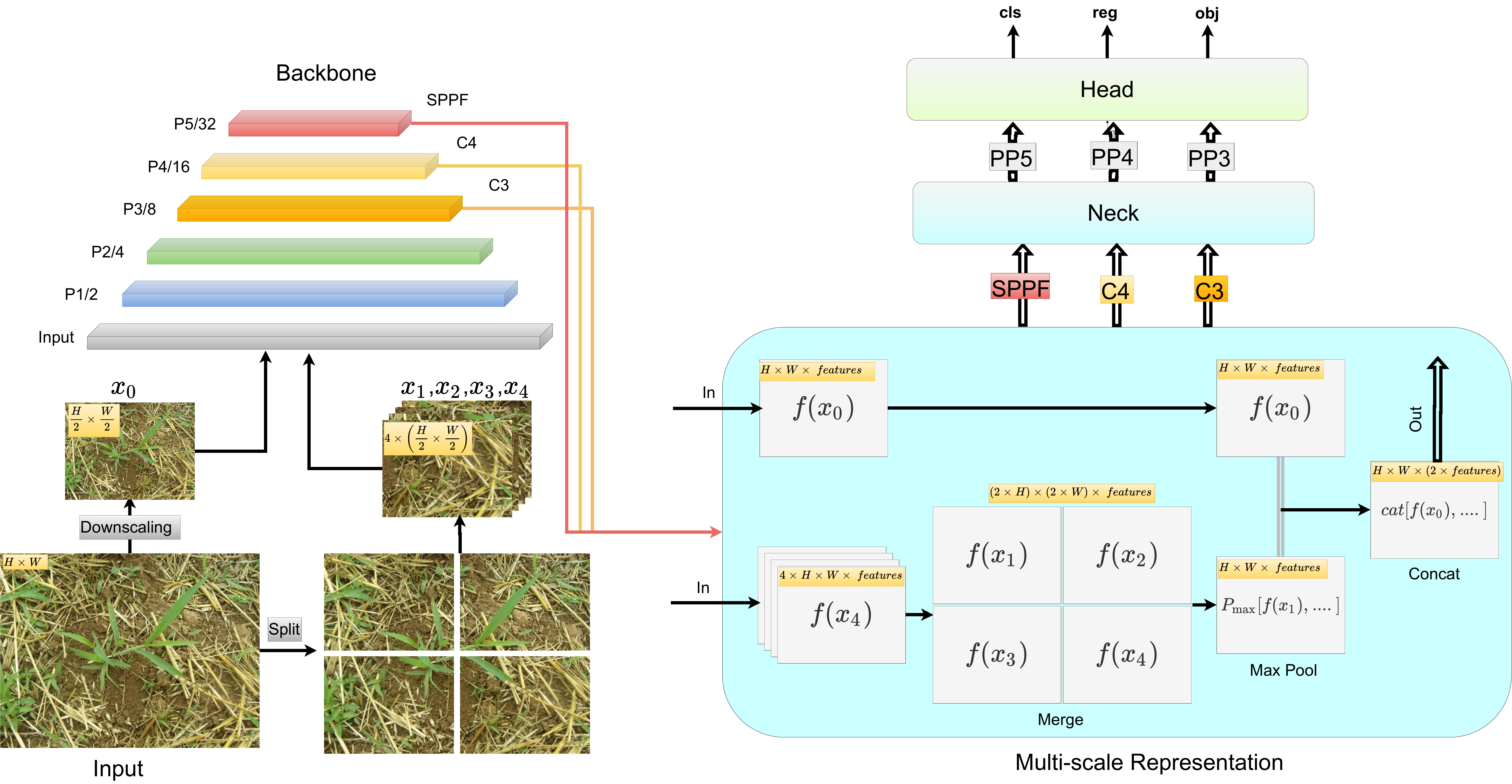}
    \caption{Multi-scale feature extraction and fusion in the YOLOv5 detector. The input image is first interpolated to different scales (e.g., $640^2$ and $1280^2$) and split into several sub-images of the same size as the default input size ($1280^2 \rightarrow 4 \times 640^2$). For each scale, all sub-images are fed into the same backbone model, and the outputs (e.g., $4 \times f_x^2$) are merged into a feature map of the whole image ($2 \times f_x^2$). Feature maps of different scales are max-pooled to the original spatial size ($f_x^2$) and concatenated together. The final multi-scale feature has the same spatial shape as the single-scale feature while having a higher channel dimension (e.g., 1024 vs. 512). This process is repeated for the feature maps from the backbone (P5/32, P4/16, and P3/8),  and then fuses these features through the feature fusion network (neck) to finally generate three feature maps PP3, PP4, and PP5. After the three feature maps PP3, PP4, and PP5 were sent to the prediction head (head), the confidence calculation and bounding-box regression were executed for each pixel in the feature map using the preset prior anchor, so as to obtain a multi-dimensional array (BBoxes) including object class, class confidence, box coordinates, width, and height information.}
    \label{l_fig_8}
\end{figure}

\subsection{Multi-Scale Representation} 
\label{sec:multi-scale-Representation}

A fundamental strength of our YOLOv5-based weed detection model lies in its ability to capture features at multiple scales. This capability is essential for accurately detecting weeds of varying sizes and growth stages, which can differ significantly due to species and environmental factors. Inspired by the concept of Scaling on Scales \cite{Shi2024WhenModels}, we implement a multi-scale feature extraction strategy within the YOLOv5 framework. 

The YOLOv5 architecture typically comprises a backbone network for feature extraction and a head network for generating predictions. Our approach focuses on incorporating multi-scale features within the backbone network. During preprocessing, the input image undergoes multi-scale transformation. As shown in \cref{l_fig_8}, first, it is interpolated to different resolutions, such as a lower resolution (e.g., 640x640 pixels) and the original size, a higher resolution (e.g., 1280x1280 pixels). For the higher resolution image, a process of tiling is employed. This involves dividing the image into smaller sub-images with the same size as the model's default input (e.g., four sub-images of 640x640 pixels each).

Each scale, including both the original image and the sub-images, is then independently fed through the YOLOv5 backbone network for feature extraction. This results in multiple feature maps, one for each scale (e.g., four $f_x^2$ feature maps for the sub-images from the higher resolution).
Subsequently, the outputs from each scale are merged into a single feature map representing the entire image. Max-pooling is used to ensure all feature maps have the same spatial size ($f_x^2$). Feature maps from different scales are then concatenated along the channel dimension. This process creates a richer feature representation that incorporates information from both low-resolution (capturing fine details) and high-resolution (capturing broader context) scales. The resulting multi-scale feature has the same spatial size as a single-scale feature but with a higher channel dimension (e.g., 1024 channels compared to 512).

This process of multi-scale feature extraction and fusion is repeated for feature maps obtained from different stages of the backbone network (e.g., P5/32, P4/16, and P3/8 in standard YOLOv5). These features are then further processed through the feature fusion network (neck) to generate three final feature maps (PP3, PP4, and PP5).
The final concatenated multi-scale features (PP3, PP4, and PP5) are fed into the YOLOv5 prediction head (head). This head network processes these features to perform confidence calculation and bounding-box regression for each pixel in the feature map using pre-defined anchor boxes. This results in a multi-dimensional array (BBoxes) containing information such as object class, class confidence, bounding box coordinates, width, and height for the detected weeds in the image.

By incorporating information from various scales, this multi-scale approach allows the model to effectively detect weeds regardless of their size or growth stage. It achieves this by capturing both fine-grained details for smaller weeds and broader context for larger weeds, leading to more robust and accurate weed detection in real-world scenarios.

 \begin{figure}[!t]
    \centering
    \includegraphics[width=0.90\linewidth]{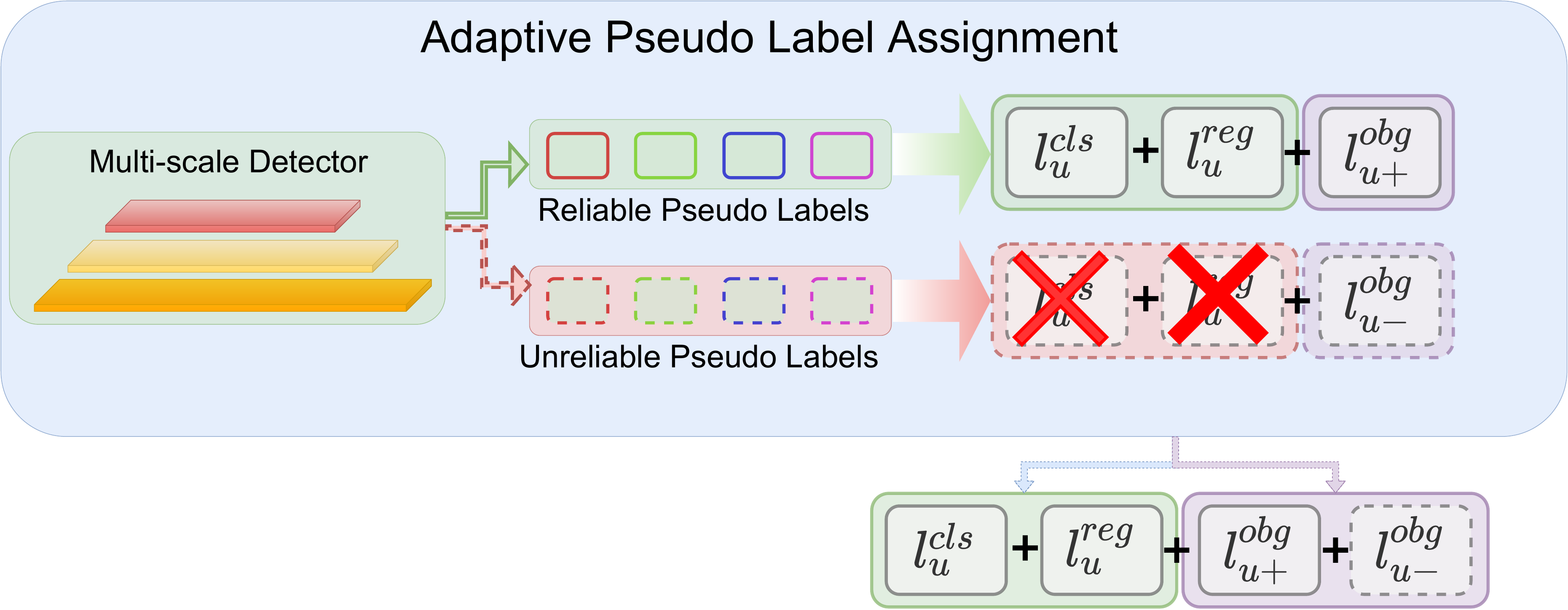}
    \caption{The Adaptive Pseudo Label Assignment process in Semi-Supervised Object Detection (SSOD). The Multi-Scale Detector analyzes data and assigns pseudo labels, which are categorized as reliable or unreliable based on two thresholds. Unreliable labels are handled specifically to improve the model's performance. The overall loss function combines supervised and unsupervised loss, with a weighting factor to balance their contributions. Indicator functions are used to selectively apply loss components based on the pseudo label score. This process effectively leverages both labeled and unlabeled data, improving model performance by handling the inherent uncertainty in pseudo labels.}
    \label{l_fig_10}
\end{figure}

\subsection{Adaptive Pseudo Label Assignment}
To leverage the unlabeled data, we introduce an adaptive pseudo label assignment process. This process involves generating pseudo labels for the unlabeled data based on the current model predictions. 
One of the core challenges in Semi-Supervised Object Detection (SSOD) is assigning accurate pseudo labels to unlabeled data. Suboptimal assignments can introduce inconsistencies and negatively impact the overall performance of the mutual learning mechanism. 

A common strategy, Pseudo Label Filter, removes labels with scores below a set threshold. This approach aims for speed but has limitations.  For example, using a low threshold can lead to including incorrect labels, while a high threshold could exclude valuable ones. This can hinder the training process and prevent the network from converging effectively.

Here, we introduce the Adaptive Pseudo Label Assigner, a method that offers a more refined approach to assigning pseudo labels generated by the Multi-Scale Detector model.
Adaptive Pseudo Label Assigner categorizes pseudo labels into two groups: reliable and unreliable. High and low thresholds ( ${\tau_{1},\tau_{2}}$ ) are used to distinguish them. Labels with scores falling between these thresholds are considered unreliable. Ignoring the loss associated with these unreliable labels directly improves the performance of the Multi-Scale Detector.
% , as shown in Table~\ref{tab:pseudo label assigner}.  

Beyond addressing the limitations of Pseudo Label Filter, Adaptive Pseudo Label Assigner incorporates an unsupervised loss function that effectively leverages these unreliable labels.  The overall loss function in SSOD using Multi-Scale Detector is defined as the sum of a supervised loss and an unsupervised loss, with a weighting factor ( $\lambda$ ) to balance their contributions (Equation~\ref{eq:total}).
\begin{equation}
  L = L_s + \lambda L_u
  \label{eq:total}
\end{equation}

\noindent
The supervised loss ( $L_s$ ) follows the standard formulation used in previous work (Equation~\ref{eq:supervised}).  
\begin{equation}
\begin{aligned}
L_s & = \sum_{h,w}(CE(X_{(h,w)}^{cls}, Y_{(h,w)}^{cls}) + CIoU(X_{(h,w)}^{reg}, Y_{(h,w)}^{reg})  & + CE(X_{(h,w)}^{obj}, Y_{h,w}^{obj})),
\end{aligned}
\label{eq:supervised}
\end{equation}
where,
 $CE(X_{(h,w)}^{cls}, Y_{(h,w)}^{cls})$: Classification loss (cross-entropy) between the predicted classification probabilities from the model, $X_{(h,w)}^{cls}$ (at location $(h,w)$ on the feature map), and the ground truth class labels, $Y_{(h,w)}^{cls}$.
 $CIoU(X_{(h,w)}^{reg}, Y_{(h,w)}^{reg})$: Complete Intersection over Union (CIoU) loss, measuring the distance between the predicted bounding boxes from the model, $X_{(h,w)}^{reg}$, and the ground truth bounding boxes, $Y_{(h,w)}^{reg}$ (at location $(h,w)$).
 $CE(X_{(h,w)}^{obj}, Y_{h,w}^{obj})$: Objectness loss (cross-entropy) between the predicted objectness score from the model, $X_{(h,w)}^{obj}$, and the ground truth objectness labels, $Y_{h,w}^{obj}$ (at location $(h,w)$). Objectness indicates the presence or absence of an object in a specific location.

\noindent
The unsupervised loss ( $L_u$ ) is further broken down into classification loss ( $L_u^{cls}$ ), regression loss ( $L_u^{reg}$ ), and objectness loss ( $L_u^{obj}$ ) (Equations~\ref{eq:cls_u} - \ref{eq:obj_u}).
\begin{equation}
\begin{aligned}
L_u = L^{cls}_u + L^{reg}_u + L^{obj}_u
\end{aligned}
\label{eq:L_u}
\end{equation}

\begin{equation}
\begin{aligned}
L^{cls}_u &= \sum_{h,w}(\mathbb{1}_{\{p_{(h,w)}>=\tau_{2}\}}CE(X_{(h,w)}^{cls}, \hat{Y}_{(h,w)}^{cls})),
\end{aligned}
\label{eq:cls_u}
\end{equation}
where,
 $L^{cls}_u$: Classification loss for unreliable pseudo labels.
 $\sum_{h,w}$: Summation over height $(h)$ and width $(w)$ of the feature map.
 $\mathbb{1}_{\{p_{(h,w)}>=\tau_{2}\}}$: Indicator function that outputs 1 if the pseudo label score at location $(h,w)$ on the feature map, denoted by $p_{(h,w)}$, is greater than or equal to the high threshold $\tau_2$. This ensures the loss is only applied to unreliable pseudo labels with high classification scores.
 $CE(X_{(h,w)}^{cls}, \hat{Y}_{(h,w)}^{cls})$: Cross-entropy loss function. It measures the difference between the predicted classification probabilities from the model, $X_{(h,w)}^{cls}$, and the soft label (explained below), $\hat{Y}_{(h,w)}^{cls}$, at location $(h,w)$.
 $\hat{Y}_{(h,w)}^{cls}$: Soft label for classification. Unlike a standard one-hot encoded label, a soft label represents the probability distribution of the class membership for a particular location. In this case, it avoids assigning these unreliable pseudo labels definitively as background or positive samples.

% \begin{small}
\begin{equation}
\begin{aligned}
L^{reg}_u = \sum_{h,w}(\mathbb{1}_{\{p_{(h,w)}>=\tau_{2} \text{ or } \hat{obj}_{(h,w)} > 0.99 \}}CIoU(X_{(h,w)}^{reg}, \hat{Y}_{(h,w)}^{reg})),
\label{eq:reg_u}
\end{aligned}
\end{equation}
% \end{small}
where,
 $L^{reg}_u$: Regression loss for unreliable pseudo labels.
 $\sum_{h,w}$: Summation over height $(h)$ and width $(w)$ of the feature map.
 $\mathbb{1}_{\{p_{(h,w)}>=\tau_{2} \text{ or } \hat{obj}_{(h,w)} > 0.99\}}:$ Indicator function that outputs 1 if either condition is met:
    1) The pseudo label score at location $(h,w)$ on the feature map, denoted by $p_{(h,w)}$, is greater than or equal to the high threshold $\tau_2$ (similar to the classification loss).
    or 2) The objectness score, $\hat{obj}_{(h,w)}$, at location $(h,w)$ is greater than 0.99. This captures unreliable labels with potentially inaccurate bounding boxes but high confidence in containing an object.
 $CIoU(X_{(h,w)}^{reg}, \hat{Y}_{(h,w)}^{reg})$: Complete Intersection over Union (CIoU) loss function. It measures the distance between the predicted bounding box from the model, $X_{(h,w)}^{reg}$, and the predicted bounding box for the ground truth, $\hat{Y}_{(h,w)}^{reg}$, at location $(h,w)$.

\begin{small}
\begin{equation}
\begin{aligned}
L^{obj}_u = \sum_{h,w}(\mathbb{1}_{\{p_{(h,w)}<=\tau_{1}\}}CE(X_{(h,w)}^{obj}, \textbf{0}) + \mathbb{1}_{\{p_{(h,w)}>=\tau_{2}\}}CE(X_{(h,w)}^{obj}, \hat{Y}_{(h,w)}^{obj}))  + \mathbb{1}_{\{\tau_{1}< p_{(h,w)} <\tau_{2}\}}CE(X_{(h,w)}^{obj}, \hat{obj}_{(h,w)}) ),
\end{aligned}
\label{eq:obj_u}
\end{equation}
\end{small}
where,
 $L^{obj}_u$: Objectness loss for unreliable pseudo labels.
 $\sum_{h,w}$: Summation over height $(h)$ and width $(w)$ of the feature map.
 $\mathbb{1}_{\{p_{(h,w)}<=\tau_{1}\}}$: Indicator function that outputs 1 if the pseudo label score at location $(h,w)$ on the feature map, denoted by $p_{(h,w)}$, is less than or equal to the low threshold $\tau_1$. This targets low-confidence pseudo labels.
 $CE(X_{(h,w)}^{obj}, \textbf{0})$: Cross-entropy loss function. Here, it calculates the loss between the predicted objectness score from the model, $X_{(h,w)}^{obj}$, and a target of 0. This enforces the model to assign these low-confidence labels as background.
 $\mathbb{1}_{\{p_{(h,w)}>=\tau_{2}\}}$: Indicator function that outputs 1 if the pseudo label score at location $(h,w)$ is greater than or equal to the high threshold $\tau_2$. This targets highly confident pseudo labels. 
 $CE(X_{(h,w)}^{obj}, \hat{Y}_{(h,w)}^{obj})$: Cross-entropy loss function. It measures the difference between the predicted objectness score, $X_{(h,w)}^{obj}$, and the predicted objectness score for the ground truth, $\hat{Y}_{(h,w)}^{obj}$, at location $(h,w)$. This ensures these highly confident pseudo labels maintain their objectness.
 $\mathbb{1}_{\{\tau_{1}< p_{(h,w)} <\tau_{2}\}}$: Indicator function that outputs 1 if the pseudo label score at location $(h,w)$ falls within the unreliable range (between the low and high thresholds, $\tau_1$ and $\tau_2$). This targets unreliable pseudo labels.
 $CE(X_{(h,w)}^{obj}, \hat{obj}_{(h,w)})$: Cross-entropy loss function. It calculates the loss between the predicted objectness score, $X_{(h,w)}^{obj}$, and a soft label for objectness, $\hat{obj}_{(h,w)}$. This allows the model to learn from these unreliable labels without strong biases towards background or object classes.

Adaptive Pseudo Label Assigner leverages indicator functions ( $\mathbb{1}_{\{\cdot\}}$ ) to selectively apply loss components based on the pseudo label score ( $p_{(h,w)}$ ).  
For highly reliable pseudo labels (scores above ${\tau_2}$ ), all loss terms are calculated.  For unreliable pseudo labels with high classification scores, only the objectness loss is considered, as shown in \cref{l_fig_10}.  
The target for the classification loss is replaced with a soft label ( $\hat{obj}_{h,w}$ ) that avoids assigning these labels as background or positive samples.  Finally, for unreliable labels with high objectness scores (greater than 0.99), the regression loss is computed.  

% Our observations during SSOD training on the COCO dataset indicate that over 70\% of unreliable pseudo labels are actually false positives due to inaccurate bounding boxes. Only a small percentage (around 6\%) have incorrect classifications with precise bounding boxes.  Therefore, Adaptive Pseudo Label Assigner prioritizes improving the regression accuracy of unreliable labels using $L_u^{reg}$ to convert them into true positives. 

% This approach with Adaptive Pseudo Label Assigner effectively suppresses inconsistencies in pseudo labels through a soft label learning mechanism, without affecting the contribution of reliable pseudo labels.  In thee following section, ................
% For a deeper dive into the details, please refer to the Appendix.

\subsection{Multi-Scale Detector}

In this work, we employ a multi-scale detector that leverages a mean teacher framework \cite{Deng2021UnbiasedDetection} in conjunction with our previously described Multi-Scale Representation module. The mean teacher framework is a mutual learning mechanism that drives a "teacher" model and a "student" model to iteratively improve each other. Our approach follows this principle, where the teacher model generates pseudo labels for unlabeled data, and the student model is trained on both labeled and pseudo-labeled data.

The foundation of our multi-scale detector is YOLOv5 \cite{Jocher2022Ultralytics/yolov5:Inference}, a widely-used one-stage anchor-based detector in the industry. We chose YOLOv5 due to its popularity, deployment practicality, and rapid training speed. The overall pipeline of our framework is illustrated in Figure~\ref{l_fig_9}.
During each training iteration, the following steps are performed:
1) Student Learns from Labeled and Pseudo-Labeled Data: The student model first trained on both the labeled data and the strongly-augmented unlabeled data with the corresponding pseudo labels.
2) Teacher Generates Pseudo Labels: The teacher model is then  generates pseudo labels on the weakly-augmented unlabeled data. These pseudo labels act as supervision signals for the corresponding strongly-augmented version of the unlabeled data.
Throughout the training process, the teacher model gradually updates its weights from the student model using an exponential moving average (EMA) strategy.

%%%%%%%%%%%%%%%%%%%%% from %%%%%%%%%%%%%%%%%
% While the mean teacher framework addresses the challenge of limited labeled data in semi-supervised object detection (SSOD), it can still face issues related to training stability and efficiency. To mitigate these challenges, we introduce the Epoch Corresponding  method, which leverages both domain adaptation and distribution adaptation techniques to enable rapid and stable SSOD training. Our approach aims to narrow the distribution gap between labeled and unlabeled data, while also dynamically estimating the threshold values for pseudo labels at each epoch.
The mean teacher framework, although addressing the scarcity of labeled data in semi-supervised object detection (SSOD), can encounter challenges related to training stability and efficiency. To tackle these issues, we propose the Epoch Corresponding method, which synergistically combines domain adaptation and distribution adaptation techniques. Our approach aims to bridge the distribution gap between labeled and unlabeled data seamlessly. Additionally, it dynamically estimates the threshold values for pseudo labels at each epoch, enabling rapid and stable SSOD training. This dual-pronged strategy not only narrows the distributional discrepancy but also ensures a robust and efficient learning process, ultimately enhancing the performance of semi-supervised object detection models.

% As shown in Figure~\ref{l_fig_9}, in contrast to alternating and original joint training schemes, Epoch Corresponding enables the neural network to receive both labeled and unlabeled data during the initial training phase, known as the Burn-In phase. Here, domain adaptation techniques with a classifier are employed to increase the detector's capacity to discriminate between the two types of data. This effectively mitigates the overfitting effect that is commonly observed in the original approach, which relies solely on labeled data during the Burn-In phase.
Unlike alternating and original joint training schemes, the proposed Epoch Corresponding method allows the neural network to process both labeled and unlabeled data from the outset, during the initial training phase known as the Burn-In phase. During this phase, domain adaptation techniques coupled with a classifier are employed to enhance the detector's ability to differentiate between the two types of data. This approach effectively mitigates the overfitting issue that typically plagues the original method, which relies solely on labeled data during the Burn-In phase. As illustrated in Figure~\ref{l_fig_9}, by exposing the network to both labeled and unlabeled data from the beginning, while simultaneously leveraging domain adaptation techniques, Epoch Corresponding enables a more robust and comprehensive training process, reducing the risk of overfitting and improving the overall performance of the object detection model.

% The domain adaptation loss function is formulated as follows:
% \begin{equation}
% L_{da} = -\sum_{h,w}\Big[D\log p_{(h,w)} + (1-D)\log(1- p_{(h,w)})\Big],
% \end{equation}
% where $p_{(h,w)}$ is the output of the domain classifier, $D = 0$ for labeled data, and $D = 1$ for unlabeled data. We utilize the gradient reversal layer (GRL) \cite{Ganin2015UnsupervisedBackpropagation}, where the ordinary gradient descent is applied for training the domain classifier, and the sign of the gradient is reversed when passing through the GRL layer to optimize the base network.
The loss function for domain adaptation can be expressed as follows:
\begin{equation}
L_{da} = -\sum_{h,w}\Big[D\log p_{(h,w)} + (1-D)\log(1- p_{(h,w)})\Big],
\end{equation}
where, $p_{(h,w)}$ represents the output from the domain classifier. The variable $D$ is set to 0 for labeled data and 1 for unlabeled data. 
We make use of the gradient reversal layer (GRL) \cite{Ganin2015UnsupervisedBackpropagation}. GRL involves the application of standard gradient descent for the training of the domain classifier. However, when optimizing the base network, the sign of the gradient is reversed as it passes through the GRL layer.

In the initial training phase, the supervised loss for a single image can be reformulated as follows:
% \begin{small}
\begin{equation}
\begin{aligned}
L_s = \sum_{h,w}(&CE(X_{(h,w)}^{cls}, Y_{(h,w)}^{cls}) + CIoU(X_{(h,w)}^{reg}, Y_{(h,w)}^{reg})   & + CE(X_{(h,w)}^{obj} + Y_{h,w}^{obj})) + \lambda L_{da}
\end{aligned}
\end{equation}
% \end{small}
\noindent where $\lambda$ is a hyperparameter that controls the contribution of domain adaptation, set to 0.1 in our experiments. By allowing the detector to see unlabeled data during the Burn-In phase, the model's expression capability is enhanced.

Existing semi-supervised object detection (SSOD) approaches face a critical challenge when generating pseudo labels on unlabeled data during training. They require the calculation of threshold values, $\tau_1$ and $\tau_2$, for Adaptive Pseudo Label Assignment. 
While existing semi-supervised object detection (SSOD) approaches effectively leverage prior knowledge of labeled data's label distribution to compute the critical thresholds $\tau_1$ and $\tau_2$ for Adaptive Pseudo Label Assignment \cite{Chen2022LabelDetection}, this method faces limitations when applied to advanced detectors like our Multi-Scale Detector. The pivotal role played by the Mosaic data augmentation technique \cite{Xu2023EfficientYOLOv5} in such detectors disrupts the label distribution ratio, rendering the direct application of the labeled data's prior label distribution ineffective.
To overcome this challenge, we propose a distribution adaptation method inspired by the re-distribution technique introduced in LabelMatch \cite{Chen2022LabelDetection}. 
Our distribution adaptation approach dynamically assigns the thresholds $\tau_1$ and $\tau_2$ at the $k$-th epoch, accounting for the label distribution changes caused by the Mosaic data augmentation. In this method, the thresholds $\tau_1$ and $\tau_2$ at the $k$-th epoch are assigned as follows:

\begin{small}
\begin{equation}
\begin{aligned}
\tau_1^k = P_c^{k}[n_c^k\cdot \frac{N_u}{N_l}]
\end{aligned}
\end{equation}
\end{small}

\begin{small}
\begin{equation}
\begin{aligned}
\tau_2^k = P_c^{k}[\alpha\%\cdot n_c^k\cdot \frac{N_u}{N_l}],
\end{aligned}
\end{equation}
\end{small}

\noindent where the reliable ratio $\alpha$ is set to 50 for all experiments, and $P_c^{k}$ represents the list of pseudo label scores of the $c$-th class at the $k$-th epoch. 
% In the context of distribution adaptation, the symbols $N_l$ and $N_u$ are used to represent the count of labeled and unlabeled data, respectively. On the other hand, $n_c^k$ stands for the count of ground truth annotations for the $c$-th class, as calculated by the Epoch Corresponding method at the $k$-th epoch.
% By dynamically calculating the appropriate thresholds at each epoch, Epoch Corresponding enables joint training to be more adaptable to dynamic data distributions.
% The integration of domain adaptation and distribution adaptation in Epoch Corresponding effectively mitigates overfitting of neural networks to labeled data. Moreover, Epoch Corresponding dynamically estimates appropriate thresholds for pseudo labels at each epoch, achieving fast and efficient SSOD training. 
% The experimental results demonstrating these effects are presented in Section~\ref{sec:experiments}.
In the context of distribution adaptation, the symbols $N_l$ and $N_u$ represent the count of labeled and unlabeled data, respectively. On the other hand, $n_c^k$ denotes the count of ground truth annotations for the $c$-th class, as dynamically calculated by the Epoch Corresponding method at the $k$-th epoch.
By dynamically computing the appropriate thresholds at each epoch, Epoch Corresponding enables the joint training process to be more adaptable to the evolving data distributions, enhancing the overall robustness and flexibility of the semi-supervised object detection training.
The synergistic integration of domain adaptation and distribution adaptation techniques in Epoch Corresponding effectively mitigates the overfitting of neural networks to the limited labeled data. Moreover, Epoch Corresponding dynamically estimates suitable thresholds for pseudo labels at each epoch, achieving rapid and efficient semi-supervised object detection training, thereby optimizing the utilization of both labeled and unlabeled data.
The experimental results that demonstrate the efficacy of these strategies and their impact on the overall performance are presented in detail in Section~\ref{sec:experiments}.

%%%%%%%%%%%%%%%%%%%%% to %%%%%%%%%%%%%%%%%

%%%%%%%%%%%%%%%%%%%%%%%%%%%%%%%%%%%%%%%%%%%%%%%%%%%%%%%%%%%%%%%%
\section{Experiments} \label{sec:experiments}

To evaluate the effectiveness of our proposed method, we conducted a series of experiments using several agricultural datasets. The experiments were designed to assess the performance of our model in terms of weed detection accuracy and the impact of using unlabeled data and multi-scale Representation.

\subsection{Datasets}

We evaluate the performance of our method on several publicly available datasets for both object detection and domain adaptation tasks. Here, we provide a detailed breakdown of each dataset:

\noindent \textbf{  Object Detection Datasets:}
\begin{enumerate}
    \item \textbf{MS-COCO  (Microsoft Common Objects in Context) \cite{Lin2014MicrosoftContext}:} This large-scale dataset is a popular benchmark for object detection tasks. It contains images with various objects labeled with bounding boxes and corresponding class labels. We follow the experimental settings established in existing works \cite{Liu2021UnbiasedDetection,Sohn2020ADetection}. Here, we use a subset of the COCO \texttt{train2017} set for labeled training data: 1\%, 5\%, and 10\% of the total images are used as labeled data, with the remaining images serving as unlabeled data. The validation set remains the COCO \texttt{val2017} set.
% *  Agricultural Datasets: 
    \item  \textbf{CottonWeedDet12  \cite{Dang2023YOLOWeeds:Systemsb}:} This dataset specifically focuses on weed detection in cotton fields. It contains images captured under various lighting and weather conditions, with annotations for cotton plants and different weed species.
    \item \textbf{CropAndWeed  \cite{Steininger2023TheManipulation}:} This dataset features images of various agricultural crops along with co-occurring weeds. It provides a challenging scenario for weed detection due to the presence of similar-looking objects.
    \item \textbf{Palmer Amaranth  \cite{Coleman2024Multi-growthHirsutum}:} This dataset consists of images containing Palmer amaranth, a particularly difficult weed in agricultural fields. It includes images of the weed at different growth stages, providing valuable data for robust weed detection models.
    \item \textbf{RadishWheat  \cite{Rayner2022RadishWheatDatasetWeed-AI}:} This dataset focuses on weed detection in radish fields, specifically distinguishing radish plants from weeds like wild mustard. 
    \item \textbf{RoboWeedMap  \cite{Dyrmann2017RoboWeedSupportNetwork}:} This dataset is captured from robotic platforms operating in agricultural fields. It includes images with diverse lighting conditions and varying weed densities, simulating real-world weed detection scenarios.
\end{enumerate}

\noindent \textbf{  Domain Adaptation Datasets: }

    We further evaluate our method's ability to adapt to different domains using datasets from \cite{Magistri2023FromRoboticsb}:
\begin{itemize}
    \item \textbf{Sugarbeet $\rightarrow$ Sunflower: } This scenario involves adapting the model from a dataset containing sugarbeet plants to a dataset with sunflower plants. While both are agricultural crops, they have distinct visual characteristics.
    \item \textbf{Sunflower $\rightarrow$ Sugarbeet:}  This is the reverse scenario of the first, where the model adapts from sunflowers to sugarbeets.
    \item \textbf{UAV-Zurich $\rightarrow$ UAV-Bonn:}  This setting involves adapting the model from a dataset captured by a drone platform over fields in Zurich to a dataset captured by a drone over fields in Bonn. While the domains are similar (agricultural fields captured by drones), potential variations in lighting, camera characteristics, and field conditions necessitate adaptation.
    \item \textbf{UAV-Bonn $\rightarrow$ UAV-Zurich:}  This is the reverse scenario of the previous one, where the model adapts from UAV imagery captured in Bonn to Zurich.
\end{itemize}
In the domain adaptation setting, the "Source" dataset represents the domain where the model is initially trained, and the "Target" dataset represents the new domain where the model's performance is evaluated.

\subsection{Implementation Details}

% \textbf{Software and Hardware Environment:}
We developed and implemented our method using PyTorch version 1.8.
The training and testing process was conducted on a Linux platform equipped with an NVIDIA GeForce RTX 4090 GPU with 24GB of memory.

% \textbf{Training Configuration:}
 
 During training, a batch size of 16 images is used. This batch consists of an equal number (8 each) of randomly sampled images from both the labeled and unlabeled data sets. This 1:1 ratio ensures balanced learning by utilizing information from both sources effectively.
 A constant learning rate of 0.01 is employed throughout the entire training process.
 We leverage an Epoch Corresponding module that dynamically calculates the thresholds ($\tau_1$ and $\tau_2$) used for assigning pseudo labels during training with unlabeled data. This approach ensures that only high-confidence predictions from the model are used for creating pseudo labels, ultimately improving the learning process.

% \textbf{Data Augmentation:}
We employ a combination of weak and strong data augmentation techniques to enhance the model's ability to generalize to unseen data and prevent overfitting:
\begin{itemize}
    \item \textbf{Weak Augmentation: }Mosaic data augmentation is used as the primary weak augmentation technique. This method involves combining multiple images into a single one during training. Mosaic augmentation effectively increases the variety of training data and helps the model learn from diverse spatial arrangements of objects within the image.

    \item \textbf{Mixed Pseudo Labels (MixPL):} We incorporate Mixed Pseudo Labels (MixPL) \cite{Chen2023MixedDetection} during training. MixPL is a combination of Pseudo Mixup and Pseudo Mosaic, specifically designed to address the negative effects of false negative samples and supplement the training data for small and medium-sized objects within the images.

    \item \textbf{Strong Augmentation:} In addition to Mosaic augmentation, we also utilize strong augmentation techniques. These techniques include left-right flipping, random scaling, color jittering, grayscale conversion, Gaussian blur, random cutout, and color space conversion. This broader range of transformations strengthens the model's robustness by enabling it to handle variations typically encountered in real-world agricultural images.
\end{itemize}

% \textbf{Training Duration and Hyperparameter:}
The training process continues for a maximum of 500 epochs, allowing the model to learn effectively from the combined labeled and unlabeled data.
To improve model stability and reduce the impact of noise during training, we employ Exponential Moving Average (EMA) with a smoothing hyper-parameter of 0.999.

% \subsection{Baseline Comparison}
% We compared our method with several baseline models, including traditional supervised learning methods and other semi-supervised learning approaches. These models were trained using the same labeled data as our method, but without the use of unlabeled data or multi-scale Representation.

\subsection{Evaluation Metrics}

To comprehensively evaluate the performance of our proposed method and compare it to different weed detection approaches, we employ standard object detection metrics. Here's a brief explanation of each metric used:

\begin{itemize}
    \item  Average Precision (AP): This metric provides a summary of the model's overall detection quality by capturing the trade-off between precision and recall across different recall thresholds. It calculates the average of the precision values at various recall levels, providing a single score that reflects the model's ability to accurately detect weeds at varying confidence levels.

    \item  Average Precision at specific Intersection over Union (IoU) thresholds (AP$_{50:95}$): This metric assesses the model's accuracy for detecting objects with a certain overlap (IoU) between predicted and ground-truth bounding boxes. We report (AP$_{50:95}$), which is a commonly used standard for object detection tasks.   AP$_{50:95}$ specifically refers to AP averaged across IoU thresholds ranging from 50\% to 95\% (calculated every 5\%). This provides a more detailed picture of the model's ability to both identify and localize weeds precisely within the image. 

    \item  Precision (P): This metric focuses on the proportion of correctly identified weeds among all objects predicted by the model. It represents the accuracy of the model's positive detections, indicating how many of the predicted weeds are actually true positives (weeds present in the image).

    \item  Recall (R): This metric measures the completeness of the model's detections, focusing on the proportion of true positive weed detections compared to the total number of actual weeds present in the image. In simpler terms, it reflects how well the model captures all the existing weeds within an image, aiming to minimize the number of missed detections.
\end{itemize}

\subsection{Results}

This section presents the quantitative evaluation results of our proposed method for semi-supervised weed detection with multi-scale representation. We compare the performance of our approach against several baselines on various datasets.

\noindent
\textbf{Object Detection Results on COCO-standard:}

Table~\ref{tab_1} presents the experimental results on the COCO-standard dataset, evaluating different methods with varying percentages (1\%, 2\%, 5\%, and 10\%) of labeled data used for training. Our proposed method consistently achieves the highest performance across all labeled data ratios, as indicated by the bold values and blue color highlighting the improvement compared to the supervised baseline. We observe significant improvements in AP$_{50:95}$ compared to supervised and other semi-supervised baselines, demonstrating the effectiveness of our multi-scale representation technique in leveraging unlabeled data for weed detection.

Our method achieves substantial gains in AP$_{50:95}$ compared to the supervised baseline, even with a limited amount of labeled data. For example, with only 1\% of labeled data, our method achieves an AP$_{50:95}$ of 26.43, which is a significant improvement of 16.14 points compared to the supervised baseline (10.29).
The performance gap between our method and other semi-supervised baselines increases as the labeled data ratio decreases. This highlights the advantage of our multi-scale representation technique in effectively utilizing unlabeled data for weed detection. 

\noindent
\textbf{Object Detection Results on Agricultural Datasets:}

Table~\ref{tab_2} showcases the results on various agricultural datasets, including CottonWeedDet12, CropAndWeed, Palmer Amaranth, RadishWheat, and RoboWeedMap. Our proposed method consistently outperforms the supervised baseline across all datasets, as shown by the blue color highlighting the improvement in AP$_{50:95}$. These results demonstrate the generalizability of our approach to diverse agricultural scenarios.

The improvement in AP$_{50:95}$ varies depending on the dataset. This can be attributed to factors such as the inherent difficulty of weed detection in each dataset (e.g., weed size, background complexity, etc.).
Even for challenging datasets like Palmer Amaranth and RadishWheat, where the supervised baseline performance is low, our method achieves significant improvements, indicating its effectiveness in handling complex agricultural environments.

\noindent
\textbf{Domain Adaptation Results:}

Table~\ref{tab_3} presents the results on domain adaptation tasks, evaluating the model's ability to adapt to a new target domain when trained on a source domain dataset. Our method consistently improves over the supervised baseline across all four domain adaptation scenarios, as shown by the blue color highlighting the improvement. These results demonstrate the effectiveness of our approach in learning transferable features through multi-scale representation, enabling adaptation to new visual characteristics present in the target domain.

The improvement in AP metrics varies depending on the specific source and target domains. This can be attributed to the degree of visual similarity between the domains.
Our method achieves significant improvements in both precision and recall compared to the supervised baseline, indicating its ability to adapt to new domains while maintaining good detection accuracy.

The results presented in this section demonstrate the effectiveness of our proposed method for semi-supervised weed detection with multi-scale representation. Our approach consistently outperforms supervised baselines and other semi-supervised methods across various object detection and domain adaptation tasks. This highlights the potential of our method for practical applications in agricultural weed management.

% \subsection{Dataset}
% Rayner G. (2022). RadishWheatDataset. Weed-AI. Available online at: https://weed-ai.sydney.edu.au/datasets/8b8f134f-ede4-4792-b1f7-d38fc05d8127.

% Nima Teimouri (2022). RoboWeedMap. Weed-AI. Available online at: https://weed-ai.sydney.edu.au/datasets/aa0cb351-9b5a-400f-bb2e-ed02b2da3699.

% \vspace{0.3cm}
% \subsection{Experiment setup}

%  YOLOv5\cite{Jocher2022Ultralytics/yolov5:Inference}

%-------------------------------------------------------------------------
\begin{table*}[ht]
  \centering 
  \setlength{\tabcolsep}{1mm}
    \caption{Experimental results on COCO-standard ($AP_{50:95}$). All the results are the average of 5 folds.}{
   \resizebox{1.0\linewidth}{!}{

%   \resizebox*{0.97\linewidth}{\textheight-2.7cm}{
  \begin{tabular}{c|l|ccccc}
    \toprule
    \multicolumn{2}{c|} {Method} &  \%1 & \%2 & \%5 & \%10 \\ 
    \midrule
    \multirow{8}{*}{Two-stage anchor-based} & Supervised  & 9.05 & 12.70 & 18.47 & 23.86 \\ 
    & STAC\cite{Sohn2020ADetection} & 13.97 $\pm$ 0.35\small{(\textcolor{blue}{$+4.92$})} & 18.25 $\pm$ 0.25 \small{(\textcolor{blue}{$+5.91$})} & 24.38 $\pm$ 0.12  \small{(\textcolor{blue}{$+5.91$})} & 28.64 $\pm$ 0.21 \small{(\textcolor{blue}{$+4.78$})} \\ 
    & Instant Teaching\cite{Zhou2021Instant-teaching:Framework} & 18.05 $\pm$ 0.15 \small{(\textcolor{blue}{$+9.00$})} & 22.45 $\pm$ 0.15 \small{(\textcolor{blue}{$+9.75$})} & 26.75 $\pm$ 0.05 \small{(\textcolor{blue}{$+8.28$})} & 30.40 $\pm$ 0.05 \small{(\textcolor{blue}{$+6.54$})} \\ 
    & Humber teacher\cite{Tang2021HumbleDetection} & 16.96 $\pm$ 0.38 \small{(\textcolor{blue}{$+7.91$})} & 21.72 $\pm$ 0.24 \small{(\textcolor{blue}{$+9.02$})} & 27.70 $\pm$ 0.15 \small{(\textcolor{blue}{$+9.23$})} & 31.61 $\pm$ 0.28 \small{(\textcolor{blue}{$+7.75$})} \\ 
    & Unbiased Teacher\cite{Liu2021UnbiasedDetection} & 20.75 $\pm$ 0.12  \small{(\textcolor{blue}{$+11.70$})} & 24.30 $\pm$ 0.07 \small{(\textcolor{blue}{$+9.80$})} & 28.27 $\pm$ 0.11 \small{(\textcolor{blue}{$+9.80$})} & 31.50 $\pm$ 0.10 \small{(\textcolor{blue}{$+7.64$})} \\ 
    & Soft Teacher\cite{Xu2021End-to-endTeacher} & 20.46 $\pm$ 0.39 \small{(\textcolor{blue}{$+11.41$})} & - & 30.74 $\pm$ 0.08 \small{(\textcolor{blue}{$+12.27$})} & 34.04 $\pm$ 0.14  \small{(\textcolor{blue}{$+10.18$})} \\ 
    & LabelMatch\cite{Chen2022LabelDetection}&  {25.81} $\pm$ 0.28 \small{(\textcolor{blue}{$+16.76$})} & - & 32.70 $\pm$ 0.18 \small{(\textcolor{blue}{$+14.23$})} & 35.49 $\pm$ 0.17 \small{(\textcolor{blue}{$+11.63$})} \\ 
    & PseCo\cite{Li2022PseCo:Detection} & 22.43 $\pm$ 0.36 \small{(\textcolor{blue}{$+13.38$})} & 27.77 $\pm$ 0.18 \small{(\textcolor{blue}{$+15.07$})} & 32.50 $\pm$ 0.08 \small{(\textcolor{blue}{$+14.03$})} & 36.06 $\pm$ 0.24 \small{(\textcolor{blue}{$+12.20$})} \\ 
    \midrule
     
     \multirow{5}{*}{One-stage anchor-based} & Supervised & 10.29 & 13.12 & 19.28 & 24.04 \\ 
    & Unbiased Teacher \cite{Liu2021UnbiasedDetection} & 18.81 $\pm$ 0.28 \small{(\textcolor{blue}{$+9.07$})} & 22.72 $\pm$ 0.21 \small{(\textcolor{blue}{$+9.60$})} & 28.35 $\pm$ 0.12 \small{(\textcolor{blue}{$+8.15$})} & 30.34 $\pm$ 0.09 \small{(\textcolor{blue}{$+6.30$})} \\    
    & Efficient Teacher \cite{Xu2023EfficientYOLOv5}   & 23.76 $\pm$ 0.13 \small{(\textcolor{blue}{$+12.47$})} &  {28.70} $\pm$ 0.14 \small{(\textcolor{blue}{$+15.58$})} &  {34.11} $\pm$ 0.09 \small{(\textcolor{blue}{$+14.83$})}  &  {37.90} $\pm$ 0.04 \small{(\textcolor{blue}{$+13.86$})} \\ 
    
    & Ours   & \textbf{26.43} $\pm$ 0.27 \small{(\textcolor{blue}{$+16.14$})} & \textbf{29.31} $\pm$ 0.23 \small{(\textcolor{blue}{$+16.19$})} & \textbf{35.03} $\pm$ 0.14 \small{(\textcolor{blue}{$+15.75$})}  & \textbf{38.54} $\pm$ 0.09 \small{(\textcolor{blue}{$+14.50$})} \\ 

    \bottomrule
  \end{tabular}
  }}

  \label{tab_1}
\end{table*}

%%%%%%%%%%%%%%%%%%%%%%%%%%%%%%%%%%%%%%%%%%

\begin{table*}[ht]
 \centering
 \setlength{\tabcolsep}{1mm}
 \caption{Experimental results on Agricultural Datasets ($AP_{50:95}$). All the results are the average of 5 folds.}{
  \resizebox{1\linewidth}{!}{
   \begin{tabular}{l|l|c|c|c|c}
    \toprule
    % \multicolumn{1}{l|} {Dataset} & {Model} & \%1 & \%2 & \%5 & \%10 \\ 
     Dataset & Model & \%1 & \%2 & \%5 & \%10 \\ 
    \midrule
    \multirow{2}{*}{CottonWeedDet12 \cite{Dang2023YOLOWeeds:Systemsb}} 
    & Supervised & 25.6 & 42.7 & 68.8 & 73.0 \\ 
    & Ours & 39.4 $\pm$ 0.18 \small{(\textcolor{blue}{$+13.8$})} & 55.6 $\pm$ 0.29 \small{(\textcolor{blue}{$+12.9$})} & 74.3 $\pm$ 0.35 \small{(\textcolor{blue}{$+5.5$})} & 78.6 $\pm$ 0.26 \small{(\textcolor{blue}{$+5.6$})} \\ 
    \midrule
    \multirow{2}{*}{CropAndWeed \cite{Steininger2023TheManipulation}} 
    & Supervised & 30.4 & 37.5 & 44.2 & 49.7 \\ 
    & Ours & 42.8 $\pm$ 0.24 \small{(\textcolor{blue}{$+12.4$})} & 48.6 $\pm$ 0.21 \small{(\textcolor{blue}{$+11.1$})} & 52.6 $\pm$ 0.24 \small{(\textcolor{blue}{$+8.4$})} & 53.8 $\pm$ 0.31 \small{(\textcolor{blue}{$+4.1$})} \\ 
    \midrule     
    \multirow{2}{*}{Palmer amaranth \cite{Coleman2024Multi-growthHirsutum}} 
    & Supervised & 03.0 & 02.4 & 07.2 & 17.3 \\ 
    & Ours & 11.2 $\pm$ 0.22 \small{(\textcolor{blue}{$+8.2$})} & 13.4 $\pm$ 0.23 \small{(\textcolor{blue}{$+11.0$})} & 16.9 $\pm$ 0.17 \small{(\textcolor{blue}{$+9.7$})} & 22.1 $\pm$ 0.18 \small{(\textcolor{blue}{$+4.8$})} \\ 
    \midrule
    \multirow{2}{*}{RadishWheat \cite{Rayner2022RadishWheatDatasetWeed-AI}} 
    & Supervised & 06.0 & 08.8 & 18.2 & 31.2 \\ 
    & Ours & 11.3 $\pm$ 0.33 \small{(\textcolor{blue}{$+5.3$})} & 12.8 $\pm$ 0.30 \small{(\textcolor{blue}{$+4.0$})} & 21.2 $\pm$ 0.27 \small{(\textcolor{blue}{$+3.0$})} & 34.8 $\pm$ 0.26 \small{(\textcolor{blue}{$+3.6$})} \\ 
    \midrule
    \multirow{2}{*}{RoboWeedMap \cite{Dyrmann2017RoboWeedSupportNetwork}} 
    & Supervised & 02.6 & 07.4 & 13.8 & 20.6 \\ 
    & Ours & 09.3 $\pm$ 0.27 \small{(\textcolor{blue}{$+6.7$})} & 11.3 $\pm$ 0.29 \small{(\textcolor{blue}{$+3.9$})} & 20.3 $\pm$ 0.25 \small{(\textcolor{blue}{$+6.5$})} & 29.2 $\pm$ 0.16 \small{(\textcolor{blue}{$+8.6$})} \\ 
    \bottomrule
   \end{tabular}
  }
 }
 \label{tab_2}
\end{table*}
%%%%%%%%%%%%%%%%%%%%%%%%%%%%%%%%%%%%%%%%%%

%%%%%%%%%%%%%%%%%%%%%%%%%%%%%%%%%%%%%%%%%%

\begin{table*}[ht]
\centering
\setlength{\tabcolsep}{1mm}
\caption{Experimental results on different domain adaptation dataset \cite{Magistri2023FromRoboticsb}. All the results are in percentage (\%).}
\resizebox{1\linewidth}{!}{
\begin{tabular}{l|l|c|c|c|c}
\toprule
\multicolumn{2}{l|}{} & \multicolumn{4}{c}{Metric} \\
Dataset [Source $\rightarrow$ Target] & Model & P & R & AP$_{50}$ & AP$_{50:95}$ \\
\midrule
\multirow{2}{*}{Sugarbeet $\rightarrow$ Sunflower}
& Supervised & 1.90 & 28.30 & 0.90 & 0.40 \\
& Ours & 22.50 \small{(\textcolor{blue}{$+20.60$})} & 13.50 \small{(\textcolor{red}{$-14.80$})} & 9.40 \small{(\textcolor{blue}{$+8.50$})} & 5.10 \small{(\textcolor{blue}{$+4.70$})} \\
\midrule
\multirow{2}{*}{Sunflower $\rightarrow$ Sugarbeet}
& Supervised & 1.80 & 14.30 & 1.00 & 0.50 \\
& Ours & 25.80 \small{(\textcolor{blue}{$+24.00$})} & 29.20 \small{(\textcolor{blue}{$+14.90$})} & 22.50 \small{(\textcolor{blue}{$+21.50$})} & 11.80 \small{(\textcolor{blue}{$+11.30$})} \\
\midrule
\multirow{2}{*}{UAV-Zurich $\rightarrow$ UAV-Bonn}
& Supervised & 68.60 & 52.00 & 57.10 & 38.00 \\
& Ours & 76.30 \small{(\textcolor{blue}{$+7.70$})} & 61.00 \small{(\textcolor{blue}{$+9.00$})} & 64.50 \small{(\textcolor{blue}{$+7.40$})} & 44.50 \small{(\textcolor{blue}{$+6.50$})} \\
\midrule
\multirow{2}{*}{UAV-Bonn $\rightarrow$ UAV-Zurich}
& Supervised & 65.30 & 61.20 & 59.20 & 39.70 \\
& Ours & 75.00 \small{(\textcolor{blue}{$+9.70$})} & 67.90 \small{(\textcolor{blue}{$+6.70$})} & 66.90 \small{(\textcolor{blue}{$+7.70$})} & 48.20 \small{(\textcolor{blue}{$+8.50$})} \\
\bottomrule
\end{tabular}
}
\label{tab_3}
\end{table*}
%%%%%%%%%%%%%%%%%%%%%%%%%%%%%%%%%%%%%%%%%%

\subsection{Ablation Study}

In this section, we conduct ablation studies to analyze the specific contributions of various components within our proposed method. We utilize the 10\% COCO-standard dataset (one of 5 folds) for these experiments. Our focus here is to verify the specific effects of our proposed components on the widely used YOLOv5 architecture.

% \noindent
\textbf{Multi-scale Representation:}
We investigate the impact of our proposed Multi-scale Representation (MSR) module on the model's performance. Table~\ref{tab:msr} shows the results. As observed, incorporating MSR leads to an improvement in both average precision across all detection thresholds ($AP_{50:95}$) and the average precision at a threshold of 0.5 ($AP_{50}$). This suggests that MSR likely allows the model to capture features relevant to objects of different sizes. This can be crucial for accurate detection, especially for small or large objects that might otherwise be missed by a single-scale representation.

\begin{table}[h]
  \centering
  \setlength{\tabcolsep}{1mm}{
    \caption{Ablation study on Multi-scale Representation.}
    \label{tab:msr}
    \begin{tabular}{l|ll}
      \toprule
      Method & $AP_{50:95}$ & $AP_{50}$\\
      \midrule
      w/o Multi-scale Representation & 36.25 & 53.16\\
      Multi-scale Representation & \textbf{38.54} & \textbf{55.64}\\
      \bottomrule
    \end{tabular}
  }
\end{table}

% \noindent
\textbf{Adaptive Pseudo Label Assignment:}
We evaluate the effectiveness of our Adaptive Pseudo Label Assignment (APLA) strategy. Table~\ref{tab:pseudo label assigner} presents the results. Compared to supervised learning and filtering out unreliable pseudo labels, APLA achieves a significant improvement in both $AP_{50:95}$ and $AP_{50}$. 
By selectively assigning pseudo labels, APLA allows the model to leverage valuable information from the unlabeled data while minimizing the impact of potential errors. This effectively expands the training dataset and provides the model with additional learning opportunities, leading to better generalization capabilities.

\begin{table}[h]
  \centering
  \setlength{\tabcolsep}{1mm}{
    \caption{Ablation study on Adaptive Pseudo Label Assignment.}
    \label{tab:pseudo label assigner}
    \begin{tabular}{l|ll}
      \toprule
      Method & $AP_{50:95}$ & $AP_{50}$\\
      \midrule
      Supervised & 24.04 & 44.31 \\
      w/o unreliable pseudo label& 35.20& 52.00\\
      Adaptive Pseudo Label Assignment & \textbf{38.54}& \textbf{55.64} \\
      \bottomrule
    \end{tabular}
  }
\end{table}

% \noindent
\textbf{Domain Adaptation:}
Finally, we analyze the contribution of our domain adaptation technique. Table~\ref{tab:domain adaptation} shows the results. While the improvement is modest, incorporating domain adaptation leads to a slight increase in both $AP_{50:95}$ and $AP_{50}$. 
By bridging the gap between the domains, domain adaptation enabled the model to learn more transferable features and generate more precise pseudo-labels on the unlabeled data. This, in turn, facilitates a smoother and more efficient convergence during SSOD training.

\begin{table}[h]
  \centering
  \setlength{\tabcolsep}{1mm}{
    \caption{Ablation study on Domain Adaptation.}
    \label{tab:domain adaptation}
    \begin{tabular}{l|ll}
      \toprule
      Method & $AP_{50:95}$ & $AP_{50}$\\
      \midrule
      w/o domain adaptation & 38.02 & 55.10\\
      domain adaptation & \textbf{38.54} & \textbf{55.64}\\
      \bottomrule
    \end{tabular}
  }
\end{table}

%%%%%%%%%%%%%%%%%%%%%%%%%%%%%%%%%%%%%%%%%%%%%%%%%%%%%%%%%%%%%%%%

%%%%%%%%%%%%%%%%%%%%%%%%%%%%%%%%%%%%%%%%%%%%%%%%%%%%%%%%%%%%%%%%
%%%%%%%%%%%%%%%%%%%%%%%%%%%%%%%%%%%%%%%%%%%%%%%%%%%%%%%%%%%%%%%%
\section{Discussion} \label{sec:discussion}

This section discusses the key findings and future directions of our work on semi-supervised weed detection with multi-scale representation (MSR). We'll explore the benefits and limitations of our approach, while also outlining promising avenues for future research.

Our method leverages MSR to achieve robust weed detection. Key advantages include:
\begin{itemize}
    \item Improved Weed Detection Across Sizes: MSR tackles the challenge of varying weed sizes by capturing both fine-grained details and broader contextual information. This leads to more accurate detection across different growth stages.
    \item Enhanced Resilience to Variations: MSR helps the model overcome variations in image quality, lighting, and weed appearances, making it suitable for real-world agricultural environments.
    \item Efficient Feature Fusion:  Our concatenation-based approach efficiently combines features from multiple scales, maintaining a streamlined architecture for efficient learning and inference.
\end{itemize}
Ablation studies confirm the effectiveness of both MSR and the Pseudo Label Assigner in boosting weed detection performance.

% Limitations and Future Work:

While promising, there are limitations to consider:
\begin{itemize}
    \item Limited Agricultural Datasets: The availability of large, high-quality agricultural datasets for weed detection remains a challenge. Exploring data augmentation and transfer learning techniques can help address this.
    \item Computational Efficiency: While efficient, further exploration of lightweight feature extraction methods for resource-constrained devices could be beneficial.
    \item Weed Species Specificity:  Our current approach focuses on general weed detection. Future work could explore incorporating species-specific information for improved detection accuracy.
\end{itemize}

% Overall Impact and Future Directions:

Our work offers a novel approach with demonstrably strong performance in weed detection. By addressing the limitations and pursuing future directions outlined above, this research can contribute to practical weed management solutions in precision agriculture. This has the potential to improve crop yields, reduce herbicide use, and promote sustainable agricultural practices.

%%%%%%%%%%%%%%%%%%%%%%%%%%%%%%%%%%%%%%%%%%%%%%%%%%

%%%%%%%%%%%%%%%%%%%%%%%%%%%%%%%%%%%%%%%%%%%%%%%%%%%%%%%%%%%%%%%%
\section{Conclusions}  \label{sec:conclusion}

This work presented a novel approach for semi-supervised weed detection with multi-scale representation. Our method leverages a deep learning architecture that extracts features from images at different scales and effectively combines them to improve weed detection performance. We employed an Epoch Corresponding to guide the learning process with unlabeled data, enabling the model to learn from a larger pool of information.

The experimental results demonstrated the effectiveness of our approach. We achieved significant improvements in weed detection accuracy compared to supervised baselines and other semi-supervised methods across various object detection and domain adaptation tasks. This highlights the potential of our method for practical applications in agricultural weed management.

Our work opens doors for further research in this domain. Future efforts can focus on addressing limitations such as limited agricultural dataset availability and computational efficiency. Additionally, exploring weed species-specific detection and incorporating techniques to handle uncertainties in unlabeled data can be valuable directions to pursue.

%%%%%%%%%%%%%%%%%%%%%%%%%%%%%%%%%%%%%%%%%%%%%%%%%%%%%%%%%%%%%%%%

\subsection{CO2 Emission Related to Experiments}
Experiments were conducted using a private infrastructure, which has a carbon efficiency of 0.432 kgCO$_2$eq/kWh. A cumulative of 850 hours of computation was performed on hardware of type RTX 2080 Ti (TDP of 250W).
Total emissions are estimated to be 91.8 kgCO$_2$eq of which 0 percent was directly offset.
Estimations were conducted using the \href{https://mlco2.github.io/impact#compute}{Machine Learning Impact calculator} presented in \citep{Lacoste2019QuantifyingLearning}.

% More in detail, in \cref{tab:results}, we compare our proposed  model with a  ResNet-18 for landmark detection.  We find that our model reduces both training time and carbon emissions by 25\%. The ResNet-18 takes about 20 hours and emits 2.16 kgCO$_2$eq, while our model takes about 15 hours and emits 1.62 kgCO$_2$eq. Carbon emissions are a major concern for training large deep-learning models. Therefore, we believe that our method is a small step towards more efficient and eco-friendly models.

% @article{lacoste2019quantifying,
%   title={Quantifying the Carbon Emissions of Machine Learning},
%   author={Lacoste, Alexandre and Luccioni, Alexandra and Schmidt, Victor and Dandres, Thomas},
%   journal={arXiv preprint arXiv:1910.09700},
%   year={2019}
% }
%%%%%%%%%%%%%%%%%%%%%%%
% Data Availability
%%%%%%%%%%%%%%%%%%%%%%%
\section*{Data Availability}
\label{sec:Data_Availability}
All the datasets used in this paper are publicly available.

%%%%%%%%%%%%%%%%%%%%%%%%
\section*{Acknowledgement}
This research is funded by the partnership between the Australian Government's Reef Trust and the Great Barrier Reef Foundation.  

% \section*{Funding}
% This project was supported by CRC project funding from the Department of Industry, Innovation and Science, Commonwealth of Australia. Mainstream Aquaculture Group partnered with James Cook University, The University of Melbourne and four commercial farming operations to deliver this project. 

%%%%%%%%%%%%%%%%%%%%%%%
% AUTHOR CONTRIBUTION
%%%%%%%%%%%%%%%%%%%%%%%
% \section*{Author Contributions}
% xxxxxx and reviewed the manuscript.

% \paragraph{Corresponding author}:\\Correspondence to alzayat.saleh@my.jcu.edu.au

%%%%%%%%%%%%%%%%%%%%%%%
% ADDITIONAL INFORMATION
%%%%%%%%%%%%%%%%%%%%%%%
\section*{Additional Information}
% To include, in this order: \textbf{Accession codes} (where applicable); \\
\textbf{Competing interests} The authors declare no competing interests.\\
% \textbf{Ethical approval} This work was conducted with the approval of 
% The corresponding author is responsible for submitting a \href{http://www.nature.com/srep/policies/index.html#competing}{competing interests statement} on behalf of all authors of the paper. This statement must be included in the submitted article file.

%%%%%%%%%%%%%%%%%%%%%%%%
% \clearpage	
% **** bibliography**** 
% \/ \/ \/ \/ 
% \ifCLASSOPTIONcaptionsoff
% \newpage
% \fi
% \IEEEtriggeratref{8}

% % references section
% \bibliographystyle{IEEEtran}
% % \bibliographystyle{IEEEtranN}
% \bibliography{references}
	
\printcredits

\bibliographystyle{cas-model2-names}

\bibliography{references}

%%%%%%%%%%%%%%%%%%%%%%%%%%%%%%%%%%%%%%%%%%%%%%%%%%%%%%%%%%%%%%%%
\end{document}